# Word Sense Disambiguation in Persian: Can AI Finally Get It Right?

Seyed Moein Ayyoubzadeh[1*] and Kourosh Shahnazari[1*]

[1*]Sharif University of Technology.

E−mail(s): smoein.ayyoubzadeh16@sharif.edu; kourosh@aut.ac.ir;

Abstract

Homograph disambiguation, the task of distinguishing between words with identical spellings but different meanings, poses a significant challenge in natural language processing. This study introduces a dataset specifically designed for Persian homograph disambiguation. We comprehensively examine various embeddings using the cosine similarity method and assess their effectiveness in downstream classification tasks. Our investigation involves training lightweight machine learning and deep learning models for homograph disambiguation and evaluating their performance using metrics such as Accuracy, Recall, and F1 Score. Our research yields three main contributions: First, we present a newly curated Persian dataset for homograph disambiguation. Second, we conduct a comparative analysis of embeddings, highlighting their utility in different contexts. Third, we benchmark a range of models, offering guidance for practitioners in selecting appropriate strategies for homograph disambiguation tasks. In summary, this study introduces a dataset, evaluates embeddings, and benchmarks models for homograph disambiguation, providing researchers and practitioners with tools to navigate this complex task.

# 1 Introduction

The quest to enhance natural language understanding has introduced numerous challenges in the field of Natural Language Processing (NLP). Central to these challenges are tasks such as word disambiguation and the clarification of sentence meanings, which are essential for realizing the full potential of language−centric applications. Word ambiguities, where words possess multiple meanings depending on the context, complicate language processing and interpretation. Addressing these challenges is critical across various domains, including machine translation, information retrieval, sentiment analysis, and question−answering systems.

The significance of homograph disambiguation in Persian is heightened by the language's lack of diacritics or vowel markings in its written form. Unlike other languages that use diacritics to distinguish between different meanings of a word, Persian often leaves words ambiguous, with a single word potentially having up to four distinct pronunciations. For instance, the Persian word ”کرم” (kerm) can be pronounced as ”کِرِم” (kérém), ”کِرم” (kérm), and ”کَرَم” (karam), each with different meanings. This complexity can cause confusion, particularly without contextual cues, making accurate disambiguation crucial for ensuring comprehension, precise translation, and effective communication in both text−based and speech−to−text systems. Therefore, developing robust mechanisms for homograph disambiguation is essential for fully leveraging Persian language processing applications.Nicolis and Klimkov [2021]

Despite advancements in NLP, word disambiguation and sentence clarity enhancement remain intricate due to the subtleties of human language. Recent progress in pre−trained language models, such as Bidirectional Encoder Representations from Transformers (BERT), has transformed the landscape of NLP tasks. BERT's ability to capture contextual semantics and syntactic structures has led to breakthroughs in text classification, named entity recognition, and machine translation. Leveraging BERT for word disambiguation and sentence clarity promises to unravel complex semantic layers, enhancing various applications with a more nuanced understanding of language. Devlin et al. [2018]

This paper explores sentence disambiguation and clarity enhancement by leveraging BERT, a transformative language model. We introduce a novel word disambiguation dataset, meticulously curated to encompass a wide array of lexical and contextual ambiguities. Through comprehensive investigation, we aim to demonstrate BERT's efficacy in resolving word ambiguities and improving sentence comprehension.

The paper is organized as follows: Section 2 provides an overview of related works in word disambiguation and sentence enhancement. Section 3 details our methodology, including the design and construction of our innovative word disambiguation dataset. Section 4 examines BERT's architecture and capabilities, highlighting its suitability for our task. Section 5 presents our experimental setup, results, and an analysis of the findings. Finally, Section 6 summarizes our contributions, discusses their implications, and suggests directions for future research.

Through this study, we aim to bridge the gap between language understanding and ambiguity resolution, contributing to the advancement of NLP applications across diverse domains.

## 2 Related Work

The challenge of homograph disambiguation in Persian has garnered significant research interest, leading to the development of various approaches and techniques. This section reviews key contributions that have addressed the complexity of Persian homographs and advanced the precision of homograph disambiguation in this linguistic context.

Early work on homograph disambiguation includes Yarowsky [1992], who proposed an unsupervised approach based on identifying and clustering salient collocations to disambiguate word senses. Lee and Ng [2002] developed a supervised learning method using contextual features such as surrounding words and part−of−speech tags, combined with a naive Bayes classifier, achieving 92.1

Subsequent research has seen the application of various supervised models using classifiers like SVMs and neural networks for Word Sense Disambiguation (WSD) and homograph disambiguation [Navigli, 2009, Taghipour and Ng, 2015, Raganato et al., 2017]. These models rely heavily on labeled training data, which can be challenging to obtain for all word senses, prompting the exploration of knowledge−based and semi−supervised approaches.

Zhong and Ng [2012] constructed a graph representation using dictionary definitions, expanded with WordNet relations, to select the sense whose definition was most related to the context, achieving 65.6

The advent of word embeddings has revolutionized NLP, with embeddings capturing semantic information and being extensively used for WSD and homograph disambiguation. Iacobacci et al. [2016] computed word embeddings for each sense based on definitions and evaluated their similarity to the context. Khaoula et al. [2022] trained sense−specific embeddings by linking WordNet senses to occurrences in a large corpus.

More advanced contextual embeddings from models such as ELMo [Peters et al., 2018], BERT [Devlin et al., 2018], and GPT−3 have been applied to WSD. These embeddings model polysemy and consider surrounding words. Wiedemann et al. [2019] demonstrated that BERT embeddings provide improvements over static Word2Vec embeddings for WSD tasks.

A Semi−Supervised Method for Persian Homograph Disambiguation − Riahi and Sedghi Riahi and Sedghi [2012] introduced a semi−supervised approach tailored for Persian homographs, leveraging a small tagged corpus and a large untagged corpus. This method combines labeled and unlabeled data to enhance disambiguation accuracy, offering promising results by effectively navigating homograph ambiguities.

Word Sense Disambiguation of Farsi Homographs Using Thesaurus and Corpus − Makki and Homayoonpoor Makki and Homayoonpoor [2008] proposed an innovative approach utilizing thesauri and corpora to disambiguate Farsi homographs. Their method extracts insights from these linguistic resources, demonstrating a practical and effective means to improve disambiguation accuracy by leveraging existing lexical knowledge and contextual information.

Word Sense Disambiguation of Persian Homographs − Jani and Pilevar Jani and Pilevar [2012] made a significant contribution by addressing the disambiguation of Persian homographs with identical written forms but distinct meanings. Their work offers insights into strategies for navigating the complexities of Persian language processing, enhancing the understanding and communication of these homographs.

These seminal works collectively highlight the multifaceted nature of Persian homograph disambiguation. Our proposed method builds upon these foundations, incorporating advancements in language processing techniques and leveraging the capabilities of ParsBert to enhance homograph disambiguation precision in Persian. By integrating insights from these pioneering studies, our research contributes to ongoing efforts to address the challenge of homograph ambiguity in the Persian language.

# 3 Methodology

In this section, we detail the comprehensive methodology employed in our research, focusing on the creation and utilization of a meticulously curated dataset designed specifically to address the challenge of homograph disambiguation in the Persian language. We provide an in−depth exposition of the fundamental attributes and salient features of the dataset, which play a pivotal role in both the training and evaluation of models dedicated to homograph disambiguation.

## 3.1 Persian Homograph Disambiguation Dataset

The cornerstone of our research is the meticulously crafted dataset tailored to the task of homograph disambiguation. This dataset comprises a diverse assortment of sentences carefully selected to contain various homograph instances. Each sentence has been precisely annotated to facilitate an in−depth exploration of the intricacies associated with homograph disambiguation. This dataset serves as an invaluable resource for advancing the comprehension and refinement of efficient disambiguation models.

### 3.1.1 Dataset Features

Our curated dataset includes several key features that enhance its utility in both the development and evaluation phases of homograph disambiguation models. Some of the salient features are:

- Homograph: Denotes the specific homograph under consideration, present within each sentence.

- Phoneme: Represents the phonetic rendition of the homograph within the sentence.
- Sentence: Contains the textual content of the sentence housing the target homograph.

To gain comprehensive insights into the structural attributes of the dataset and uncover patterns in sentence lengths, we conducted an analysis supported by visual aids. The resulting graphical representation provides an overview of our observations and discoveries.

Distribution of Sentence Lengths The graph presented in Figure 1 illustrates the distribution of sentence lengths within the dataset. The x−axis quantifies sentence length in terms of word count, while the y−axis represents the frequency of sentences in each length category. The histogram, shaded in serene blue, captures the distribution of sentence lengths across the dataset, offering an immediate and intuitive understanding of the range of sentence structures and their frequencies.

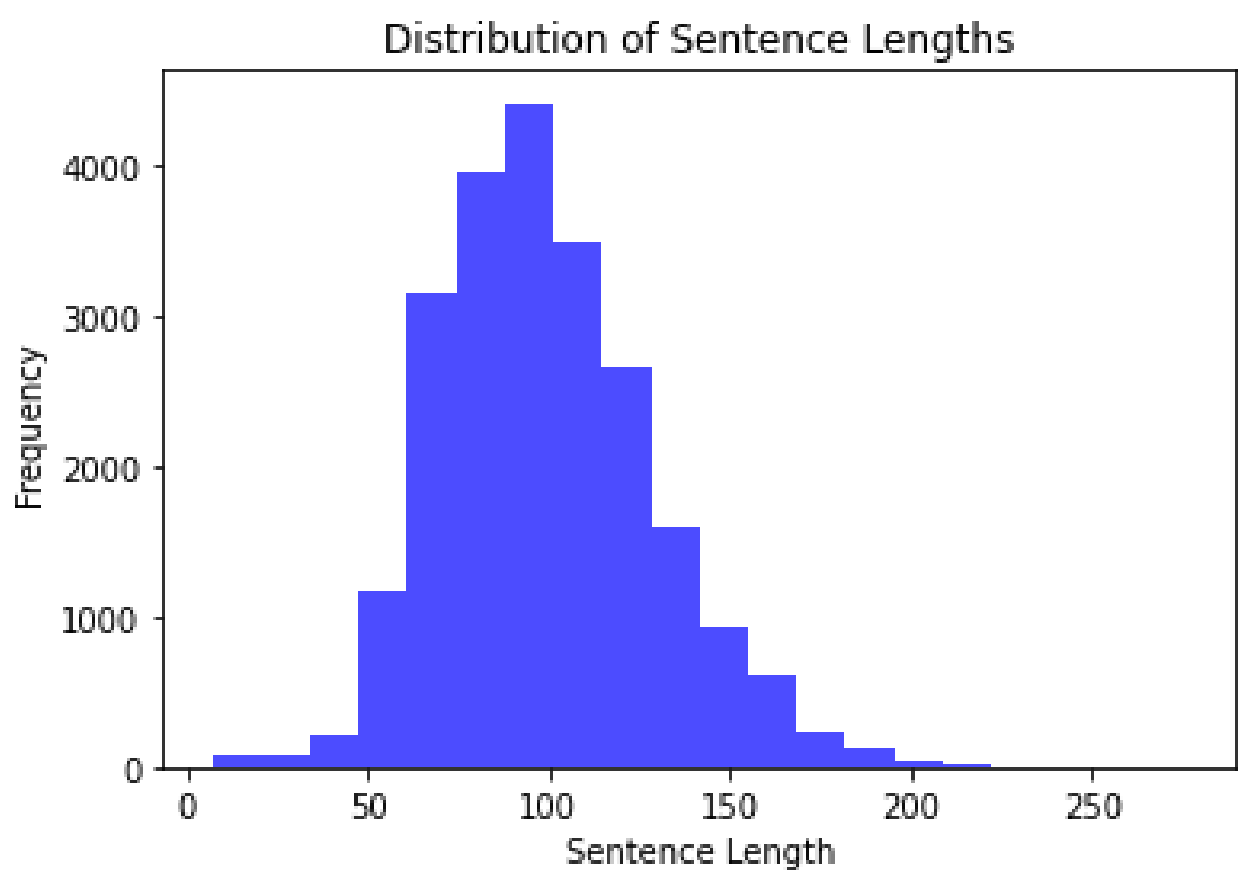

Fig. 1: Distribution of Sentence Lengths

Distribution of Homograph Positions In our examination of homograph positioning within sentences, we aimed to elucidate the occurrences and placements of homographs across the textual corpus. Figure 2 depicts the distribution of homograph positions within sentences. The x−axis signifies the position of the homograph within the tokenized sentence, while the y−axis denotes the frequency of homograph occurrences at each position. The histogram, adorned in green, encapsulates the distribution pattern, providing insights into the prevalence and positioning of homographs within the dataset's sentences.

Distribution of Homographs by Unique Phoneme Counts Our analysis of homograph distributions based on the number of unique phonemes they encompass reveals the intricacies of phonemic diversity among homographs.

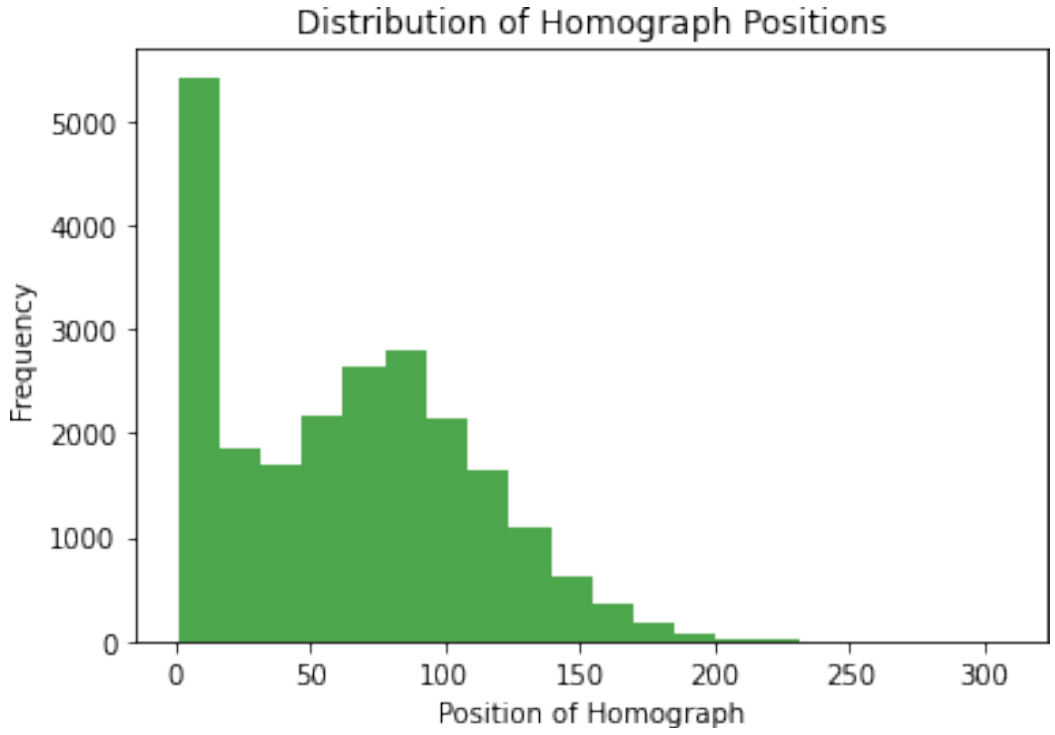

Fig. 2: Distribution of Homograph Positions

Figure 3 presents a bar plot illustrating this distribution. The x−axis denotes the number of unique phoneme counts, while the y−axis signifies the count of homographs exhibiting each specific number of unique phonemes. This visualization highlights the phonemic variety within homographs, providing insights into patterns that contribute to the nuanced landscape of our dataset.

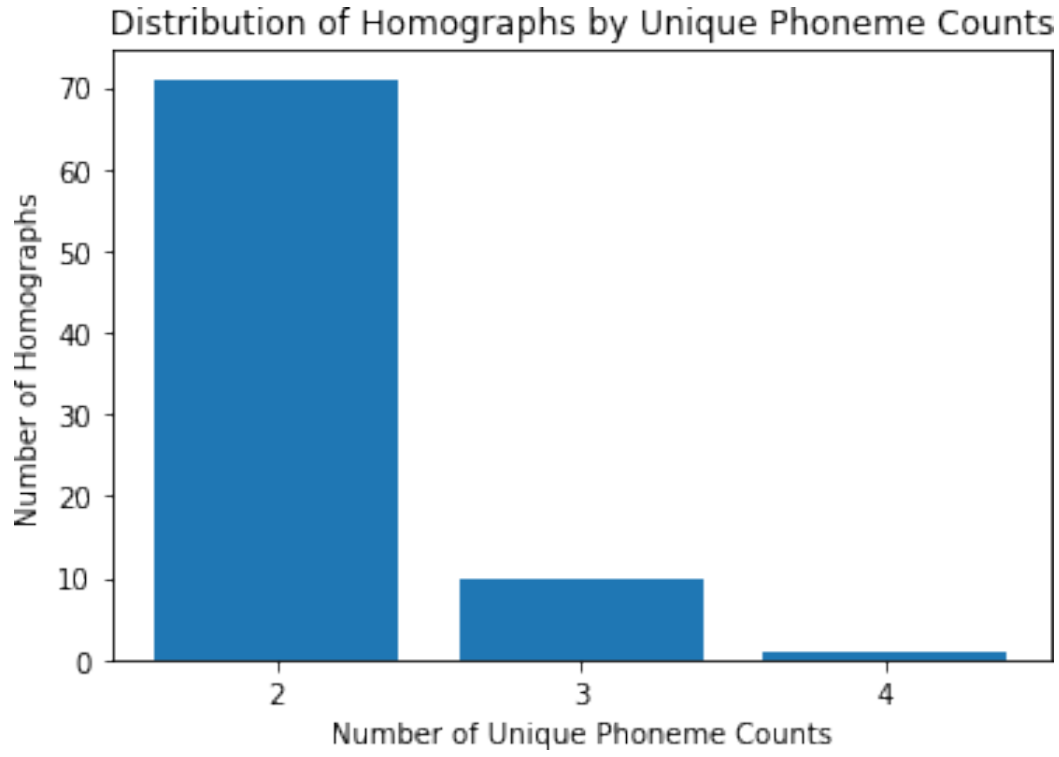

Fig. 3: Distribution of Homographs by Unique Phoneme Counts

Distribution of Homograph Lengths Additionally, we investigated the distribution of character counts in homographs within our dataset to reveal insights into the prevalence of different homograph lengths. Figure 4 showcases this distribution. The x−axis indicates the number of characters in each homograph, while the y−axis signifies the frequency of homographs within each length category. This visual representation provides a clear depiction of the distribution patterns of homograph lengths, informing the design of effective methodologies for Persian homograph disambiguation.

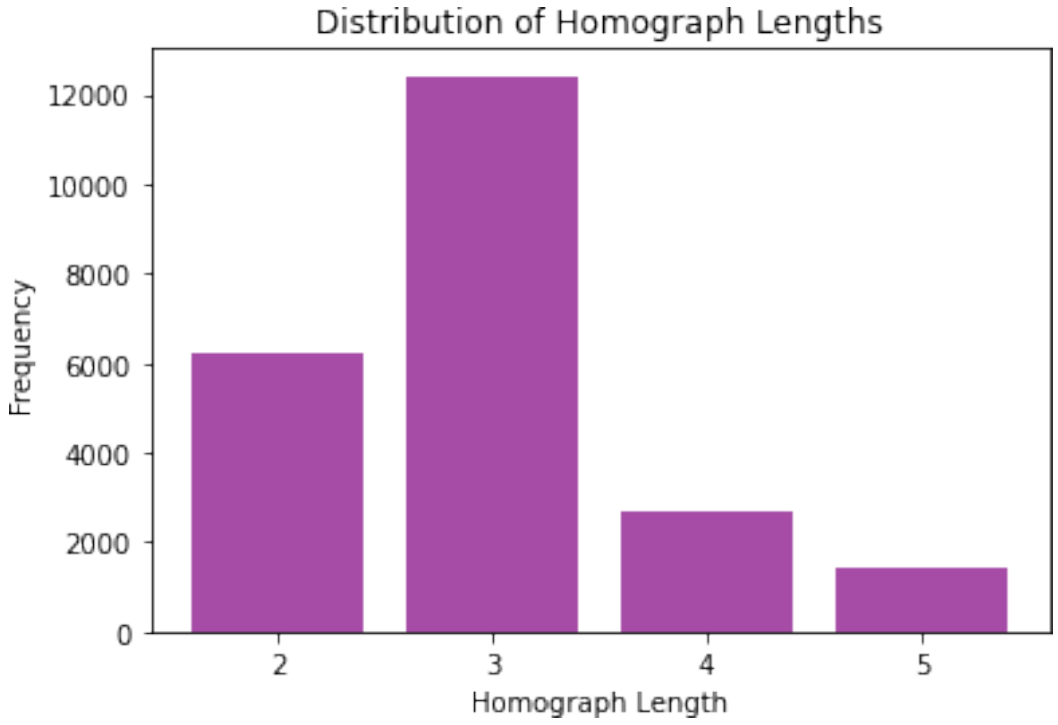

Fig. 4: Distribution of Homograph Lengths

Phonetic Diversity and Sentence Frequency Table 1 demonstrates the diversity in phonetic representations of homographs in our dataset. Each homograph is associated with multiple phonemes, capturing different pronunciations and contextual nuances. The ”Number of Sentences” column reflects the frequency of each phoneme’s occurrence in the dataset. This analysis highlights the complexity of the relationship between orthography and pronunciation, emphasizing the need for robust language understanding models to handle such linguistic intricacies.

| Homograph | Phonemes | Number of Sentences |
|---|---|---|
| اعمال | \emAl, amAl | 201, 200 |
| برنده | barande, borande | 192, 193 |
| تن | tan, ton | 193, 202 |
| جرم | jarm, jorm | 191, 199 |
| خلق | xalq, xolq | 203, 199 |
| خیر | xayyer, xeyr | 198, 192 |
| دین | deyn , din | 199, 187 |
| رب | rab, rob | 200, 204 |
| سبک | sabk, sabok | 195, 188 |
| سر | sar, ser, sor | 198, 196, 186 |
| سمت | samt, semat | 203, 201 |
| شش | SeS, SoS | 200, 194 |
| شک | Sak, Sok | 195, 183 |
| شکوه | Sekve, Sokuh | 192, 194 |
| فوت | fot, fut | 193, 196 |
| مرد | mard, mord | 198, 200 |
| مقدم | maqdam, moqaddam | 160, 202 |
| نشستن | našostan, nešastan | 197, 198 |
| کرد | kard, kord | 204, 178 |
| گزید | gazid, gozid | 200, 189 |

| Homograph | Phonemes | Number of Sentences |
|---|---|---|
| اشراف | \aSrAf, \eSrAf | 40, 45 |
| تنگ | tang, tong | 40, 43 |
| خودرو | xodro, xodru | 42, 40 |
| درک | darak, dark | 38, 43 |
| دز | dez, doz | 44, 38 |
| رحم | rahem, rahm | 40, 41 |
| سم | sam, som | 41, 39 |
| طبق | tabaq, tebq | 43, 38 |
| عرق | \araq, \erq | 42, 43 |
| غنا | qanA, qenA | 34, 40 |
| قسم | qasam, qesm | 41, 40 |
| قطر | qatar, qotr | 41, 40 |
| مبلغ | mablaq, moballeq | 41, 38 |
| مسلم | mosallam, moslem | 40, 40 |
| میل | meyl, mil | 36, 38 |
| نقل | naql , noql | 40, 42 |
| هزار | hazAr, hezAr | 39, 37 |
| پیک | peyk, pik | 43, 41 |
| کابل | kAbl, kAbol | 42, 41 |
| کشتی | keSti, koSti | 42, 34 |
| حسن | hasan, hosn | 267, 391 |
| ده | dah, deh | 404, 399 |
| سحر | sahar, sehr | 349, 400 |
| شکر | Sekar, Sokr | 403, 387 |
| عمر | \omar, \omr | 309, 394 |
| نفس | nafas, nafs | 404, 387 |
| پر | par, por | 397, 388 |
| پست | past, post | 397, 398 |
| کشت | kesht, koSt | 391, 392 |
| گل | gel, gol | 382, 399 |
| بر | bar, ber, bor | 100, 104, 94 |
| ترک | tarak, tark, tork | 95, 99, 100 |
| تو | to, tu | 105, 66 |
| جست | jast, jost | 95, 92 |
| جنگ | jang, jong | 104, 105 |
| خفت | kheft, xeffat, xoft | 101, 100, 98 |
| خم | xam, xom | 101, 97 |
| در | dar, dorr | 100, 238 |
| رفت | raft, roft | 96, 101 |
| سرور | sarvar, server, sorur | 98, 97, 99 |
| سیر | seyr, sir, siyar | 98, 100, 94 |
| صرف | sarf, serf | 104, 100 |

| Homograph | Phonemes | Number of Sentences |
|---|---|---|
| قوت | qovvat, qut | 101, 101 |
| محرم | mahram, moharram, mohrem | 100, 104, 101 |
| ملک | malak, malek, melk, molk | 97, 90, 102, 99 |
| مهر | mahr, mehr, mohr | 102, 96, 98 |
| کش | kas, kes, koS | 102, 96, 100 |
| که | key, ki | 93, 100 |
| گرده | garde, gerde, gorde | 104, 103, 100 |
| اشکال | \aSkAl, \eSkAl | 20, 20 |
| بعدی | ”badi”, ”bodi” | 20, 19 |
| رم | ram, rom | 20, 23 |
| رویه | raviyye, ruye | 21, 21 |
| سنی | senni, sonni | 20, 20 |
| عمان | ammAn, ommAn | 20, 20 |
| معین | ”moayyan”, ”moin” | 16, 20 |
| مفصل | mafsal, mofassal | 22, 21 |
| نیل | neyl, nil | 19, 21 |
| پرت | part, pert | 20, 20 |
| کنده | kande, konde | 22, 25 |
| گردان | gardAn, gordAn | 21, 20 |
| گلی | geli, goli | 20, 21 |

Table 1: Phoneme and Sentence Information

## 3.2 Dataset Preparation

We begin by preparing our dataset, which consists of a collection of homographs along with their corresponding sentences and phonetic transcriptions. Each homograph is associated with multiple pronunciations and meanings, contributing to the linguistic complexity of our dataset. We tokenize the sentences and obtain their phonetic transcriptions, forming the basis for further analysis.

## 3.3 Embedding Approach

In this section, we outline the methodology employed to generate embeddings for the homographs using the ParsBERT language model. We explore two different approaches for obtaining embeddings: utilizing the last hidden layer and calculating the average of the last four hidden layers. Our objective is to investigate the effectiveness of these methods in capturing semantic nuances and differentiating between various pronunciations of homographs.Farahani et al. [2021]

### 3.3.1 ParsBERT Language Model

ParsBERT, a variant of the BERT (Bidirectional Encoder Representations from Transformers) model, is a powerful language model pre−trained on large Persian text corpora. It serves as the foundation for our embedding generation process. We employ the pre−trained ParsBERT model to convert our input text into contextualized word embeddings, capturing intricate relationships between words in sentences. To generate embeddings for the homographs, we leverage the ParsBERT model. Specifically, we focus on two distinct methods for obtaining embeddings:

#### 3.3.1.1 Last Hidden Layer

In this approach, we extract the embeddings from the last hidden layer of the ParsBERT model. This layer encapsulates the contextual information of each token in the sentence. We use these embeddings to represent the homographs, preserving the influence of surrounding words on their meanings.

#### 3.3.1.2 Average of Last Four Hidden Layers

An alternative approach involves calculating the average of embeddings from the last four hidden layers of ParsBERT. These layers capture different levels of abstraction, ranging from syntactic to semantic information. By averaging embeddings across these layers, we aim to incorporate a broader spectrum of linguistic features.

## 3.4 Embedding Analysis

Once we generate embeddings using the two methods, we proceed with an in−depth analysis of their effectiveness. We evaluate the embeddings based on their ability to distinguish between different phonetic variations of homographs and capture their diverse meanings.To achieve this, we employed a Multi−Layer Perceptron (MLP) classifier for each homograph in our dataset. We trained and tested these classifiers to assess their performance in disambiguating homographs. Our approach involved the use of two distinct embedding methods: embeddings from the last hidden layer of ParsBert, and the average of the last four hidden layers. The methodology for assessing embeddings can be summarized as follows:

1. Classifier Training and Data Splitting: We divided the dataset into training and testing sets, using a standard test size of 0.3. Each homograph was associated with its corresponding MLP classifier.
2. Embedding Extraction: We extracted embeddings from the ParsBert model to capture the semantic information of the homographs and their contexts. Both the embeddings from the last hidden layer and the average of the last four hidden layers were utilized.
3. Categorization of Homographs: Homographs were categorized based on the count of their associated phonemes. This categorization allowed us to evaluate the impact of embedding methods on different levels of disambiguation complexity.

4. Evaluation and Result Comparison: After training the MLP classifiers, we evaluated their performance on the testing set for each category of homographs. Accuracy scores were calculated for different phoneme categories and embedding methods.

## 3.5 Evaluation Metrics

To quantify the performance of our embedding methods, we utilize several evaluation metrics. We measure the cosine similarity between embeddings corresponding to different pronunciations of the same homograph, as well as embeddings associated with distinct homographs. Additionally, we perform downstream tasks such as classification to assess the utility of the embeddings in capturing semantic information.

### 3.5.1 Interpretation

Comparing the two methods allows us to understand the strengths and weaknesses of the embeddings generated by each method:

- Accuracy of Trained MLP: Higher accuracy indicates that the embeddings effectively capture semantic relationships and features that contribute to accurate predictions. However, this method might not provide insights into the nature of semantic relationships and might not be able to differentiate subtle differences.
- Cosine Similarity: The cosine similarity method focuses on directly comparing the embeddings of pairs of words using the cosine similarity metric. Cosine similarity measures the angle between two vectors and provides a value between $-1$ and 1, where higher values indicate greater similarity. By comparing the cosine similarity values between pairs of words, we can gain insights into how well the embeddings differentiate between words. High cosine similarity values imply that embeddings are close in the vector space, suggesting similar semantic meanings. Conversely, low cosine similarity values suggest distinct semantic meanings. This method helps us understand how well embeddings distinguish between different semantic concepts. High cosine similarity values might indicate that embeddings are capturing some shared context, while low values might signify effective differentiation.

### 3.5.2 Implications

The analysis of different embeddings using these two methods helps us make informed decisions about which embeddings are more suitable for specific tasks. If the goal is to achieve high accuracy in downstream tasks, embeddings with better performance in the MLP−based evaluation might be preferred. On the other hand, if understanding the semantic relationships between words is crucial, examining cosine similarity patterns could be more insightful.

Ultimately, the choice of method depends on the specific goals of the analysis and the tasks the embeddings will be used for. A combination of both methods can provide a more comprehensive understanding of the strengths and weaknesses of the embeddings.

## 3.6 Homograph Disambiguation Methodology

In this section, we outline our proposed methodology for homograph disambiguation, which involves training different classifiers for each homograph. Our approach aims to leverage contextual embeddings to effectively distinguish between different meanings of homographs.

### 3.6.1 Classifier Training

To disambiguate the meanings of homographs, we adopt a personalized classifier approach. Specifically, we train a separate classifier for each distinct homograph present in the dataset. This allows us to capture the unique semantic nuances associated with each homograph. For each homograph, we partition the data into training and testing sets using an 80/20 split. We opted to employ a variety of lightweight machine learning and deep learning models as our base classifiers, taking advantage of their ability to efficiently learn intricate patterns from embeddings. In the training phase, the embeddings are employed as input features, while the corresponding sense labels serve as target outputs.

### 3.6.2 Data Preprocessing

We start by preprocessing the dataset, which consists of a collection of sentences containing homographs. For each homograph, we gather the associated sentences and their corresponding embeddings. These embeddings are obtained from the ParsBERT model, utilizing both the embeddings of the last hidden layer and the average of the last four hidden layers.

### 3.6.3 Evaluation and Comparison

After training, we evaluate the performance of our classifiers using the testing data. We compute various evaluation metrics, such as accuracy, precision, recall, and F1−score, to assess the effectiveness of our approach.

## 3.7 Experimental Setup

Our experiments were conducted using a computing system consisting of an Intel(R) Core(TM) i7−8750H CPU operating at 2.20GHz and an NVIDIA GeForce GTX 1050Ti GPU. This provided the necessary computational resources for our machine−learning tasks.

We utilized Python and key libraries including HuggingFace Transformers and Scikit−learn to extract embeddings and evaluate our models. The main machine learning models and their key parameters are summarized below:

- K−Nearest Neighbors (KNN): K=7 neighbors
- Multilayer Perceptron (MLP): 2 hidden layers with 100 neurons
- Random Forest: 100 estimators, Gini criterion for the quality of splits

The experiments follow a methodology to ensure unbiased analysis. The system hardware combined with Python libraries and optimized model parameters provides an ideal experimental platform to thoroughly evaluate and compare text embeddings.

# 4 Results

## 4.1 Analysis of Different Embeddings

In this analysis, we explore the effectiveness of two different methods for generating word embeddings by considering their impact on two evaluation metrics: the accuracy of a trained MLP (Multi−Layer Perceptron) and the cosine similarity between embeddings. The goal is to understand how these two methods perform in capturing semantic relationships between words.

### 4.1.1 Comparison Based on the Accuracy of Trained MLP

The results of our homograph disambiguation approach are presented in Figure 5. We evaluated the performance of two different types of embeddings: embeddings from the last hidden layer (Last Layer Embeddings) and the average of the last four hidden layers (average last Four Layers Embeddings). The accuracies of the classifiers trained using these embeddings for homograph disambiguation are compared based on the number of phonemes present in each homograph.

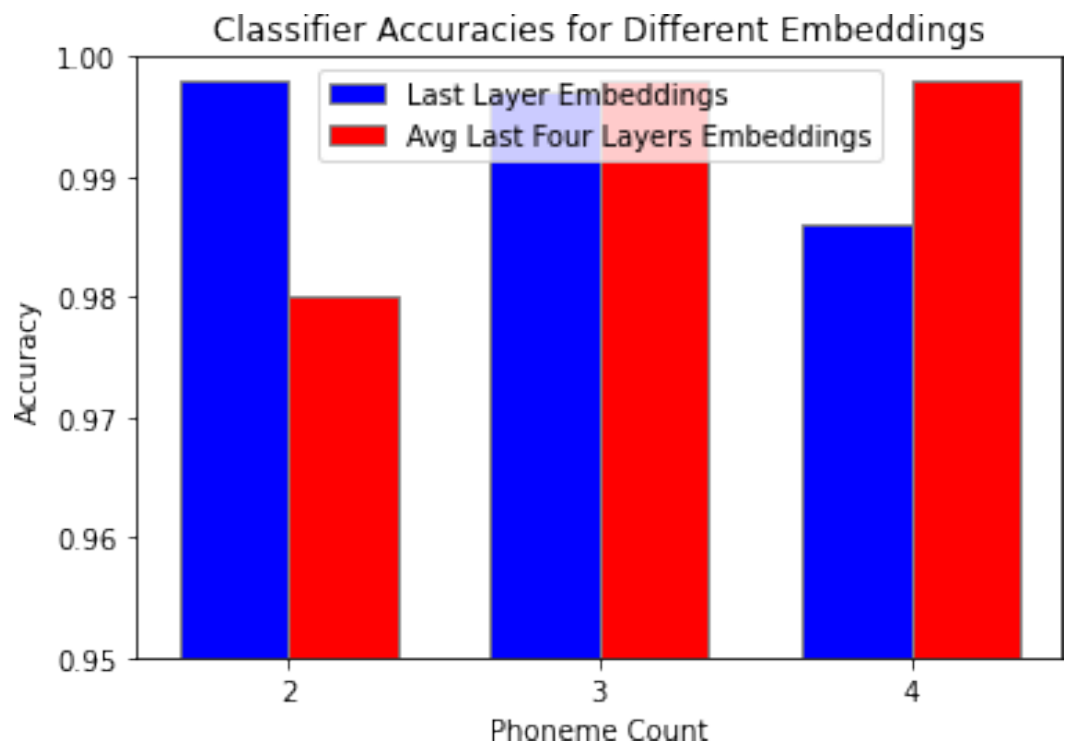

Fig. 5: Classifier Accuracies for Different Embeddings

As depicted in the figure, the accuracies vary across different homographs and phoneme counts. It is evident that for homographs with a phoneme count of 2, Last Layer Embeddings outperform Avg Last Four Layers Embeddings. For homographs with a phoneme count of 3, we observe they performed almost equally, with Avg Last Four Layers Embeddings maintaining a higher accuracy. However, as the phoneme count increases to 4, the accuracy of classifiers trained with Avg Last Four Layer Embeddings surpasses that of classifiers trained with Last Layer Embeddings.

This suggests that the choice of embeddings has an impact on the performance of homograph disambiguation, and it is influenced by the complexity of the homograph's phonemic structure. The phenomenon can be attributed to the different levels of linguistic information captured by the two types of embeddings. Last Layer Embeddings may excel in capturing fine−grained phonetic nuances, while Avg Last Four Layer Embeddings might capture broader contextual information relevant to disambiguation.

In conclusion, our results highlight the importance of selecting appropriate embeddings for homograph disambiguation, considering both the linguistic characteristics of the homographs and the structure of the neural network.

### 4.1.2 Comparison Based on Cosine Similarity

We computed the cosine similarity between pairs of embeddings for each homograph and calculated the mean cosine similarity for each embedding. Then we examined the distribution of cosine similarity values between pairs of embeddings for each homograph. Figure 6 shows a histogram of the mean cosine similarity values across homographs using the average of the last four hidden layers. Figure 7 presents a histogram of the mean cosine similarity values using only the last hidden layer. Comparing these distributions provides insight into the similarity of embeddings generated by these two methods. The embeddings derived from just the last layer exhibit a distribution shifted slightly towards higher mean cosine similarity values compared to the embeddings derived from the average of the last four layers. The histogram visualization allows us to see the overall distribution and spread of the similarity values, rather than just summary statistics.

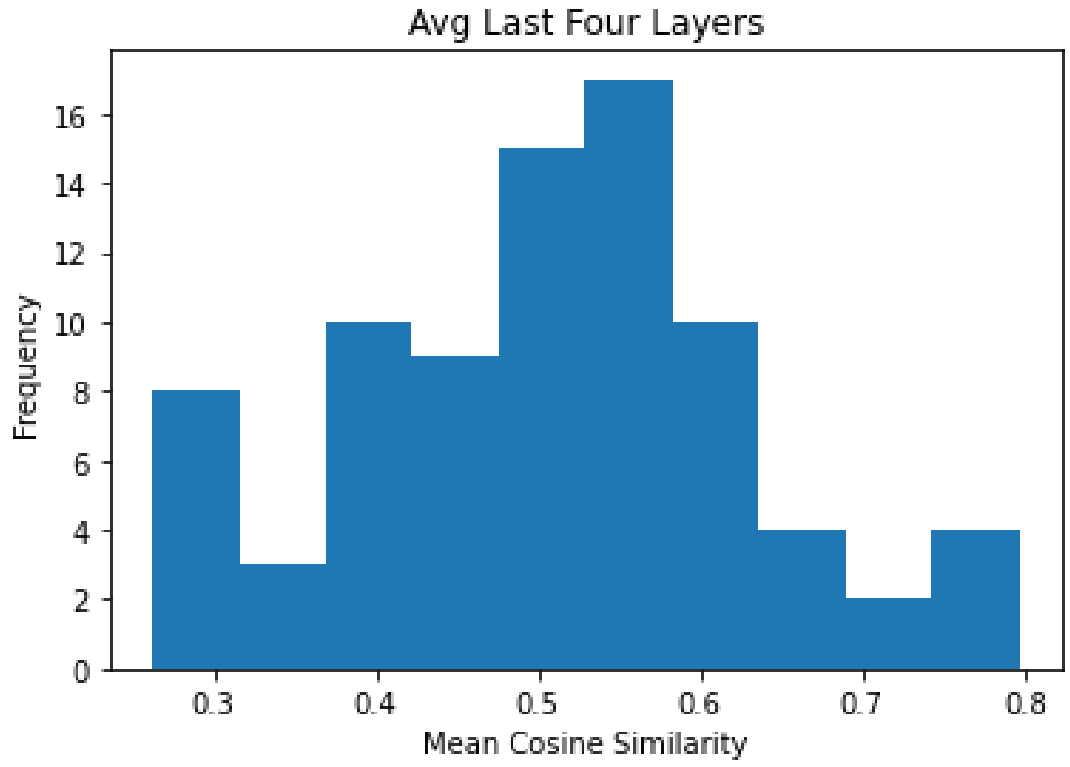

Fig. 6: Classifier Accuracies for Different Embeddings

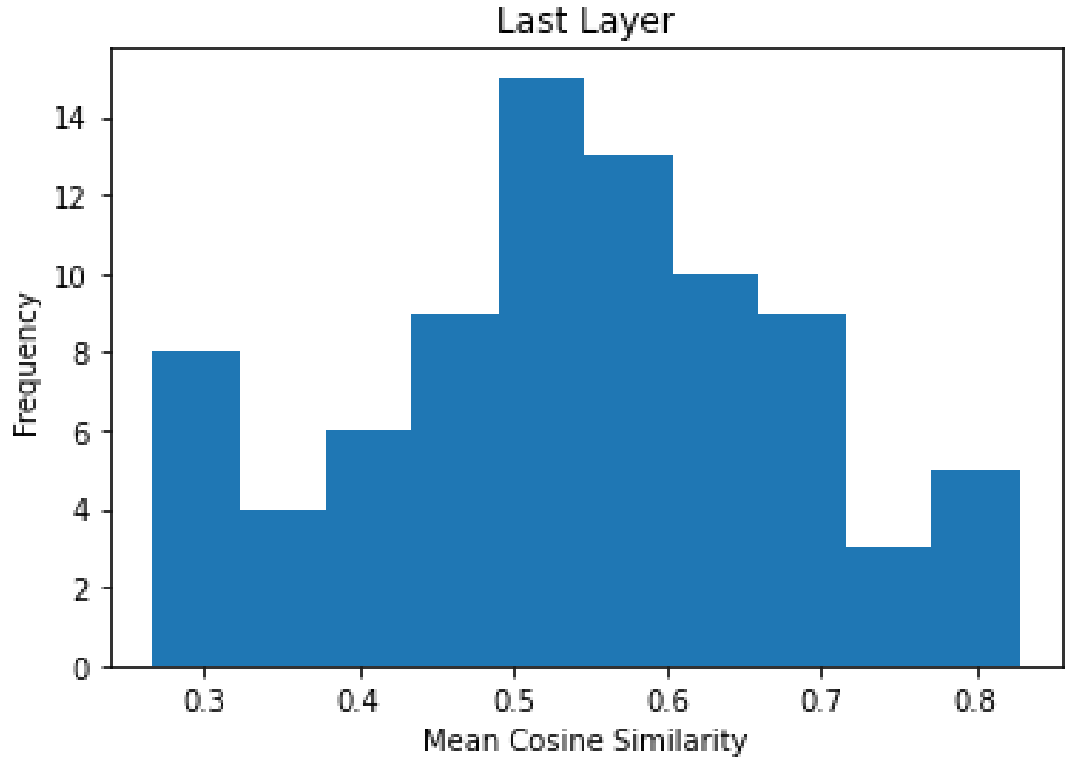

Fig. 7: Classifier Accuracies for Different Embeddings

## 4.2 Compare Different Classifiers

In this section, we present the results of our model comparison based on different evaluation metrics, including Mean Accuracy, Mean Recall, and Mean F1 Score for homograph disambiguation.

We evaluated several common machine learning classifiers on our dataset, tuning key parameters for optimal performance. The K−Nearest Neighbors (KNN) algorithm classifies samples based on a majority vote of the K closest training examples.Taud and Mas [2018] For KNN, we set K = 7 neighbors. The Multilayer Perceptron (MLP) is a feedforward artificial neural network model that uses backpropagation for training.Taud and Mas [2018] The MLP was configured with a 100−unit hidden layer, 2 hidden layers total, and sigmoid activation functions. Compared to Logistic Regression, a linear classification model, the MLP can model non−linear relationships in the data by introducing a hidden layer. Random Forest is an ensemble method that constructs multiple decision trees and aggregates their predictions.Biau and Scornet [2016] Ridge regression is a regularized linear model that handles collinearity between variables.Xingyu et al. [2022] Table 2 provides a detailed overview of the performance of these models on our dataset. Each row corresponds to a specific model's performance across the metrics considered. The tuned models allow us to effectively evaluate the tradeoffs between accuracy, recall, and F1 score on this task.

# 5 Conclusion & Future Work

In this study, we propose creating a dedicated Persian homograph disambiguation dataset to enrich the resources available in this domain. Such a dataset would empower researchers and practitioners to undertake more comprehensive

Table 2: Comparison of Model Performance using Different Metrics

| Model | Accuracy (Percent) | Recall (Percent) | F1 Score (Percent) |
|---|---|---|---|
| KNN (K = 7) | 98.94 | 98.83 | 98.82 |
| Logistic Regression | 99.70 | 99.66 | 99.68 |
| MLP (layers size = 100, layers = 2) | 99.70 | 99.66 | 99.68 |
| Random Forest (# of estimators = 100) | 99.08 | 99.02 | 99.03 |
| Ridge Classifier (alpha=1.0) | 99.61 | 99.59 | 99.59 |

investigations and contribute to the evolution of homograph disambiguation methodologies.

We analyzed homograph disambiguation using various machine−learning techniques. Our primary objective was to assess the efficacy of different models in accurately classifying homographs based on their associated phonemes. To this end, we leveraged embeddings obtained from distinct layers of a pre−trained neural network, employing them as input features for our models.

Through rigorous experimentation and meticulous evaluation, we revealed variations in the performance exhibited by the range of models. Our findings underscored the significance of selecting an appropriate model that aligns with specific objectives, as certain models showcased excellence in particular metrics. Our comparative analysis, encompassing Accuracy, Recall, and F1 Score, provided a comprehensive overview of each model's strengths.

Both the Logistic Regression and Ridge Classifier consistently delivered commendable accuracy and F1 scores, rendering them compelling choices for precise homograph disambiguation. Concurrently, the K−Nearest Neighbors model demonstrated competitive recall values, highlighting its proficiency in detecting instances of significance.

Moreover, the Multilayer Perceptron and Random Forest models exhibited well−balanced performances across diverse metrics, underscoring their versatility in handling homograph disambiguation tasks. Our evaluation not only unveiled the pivotal influence of model selection on performance but also underscored the necessity of comprehending inherent model capabilities and limitations.

In conclusion, our study contributes substantively to the homograph disambiguation field by furnishing invaluable insights into the performance intricacies of distinct machine learning models. These insights can serve as a compass for practitioners when navigating the landscape of model selection for analogous undertakings. As a future avenue of exploration, researchers could delve into more advanced embedding techniques and precision−refinement strategies to further elevate model efficacy.

# Acknowledgments

We would like to express our heartfelt appreciation to the contributors behind the remarkable tools and libraries that greatly facilitated our research journey. Specifically, we are indebted to the creators and maintainers of NumPy, Jupyter, and scikit−learn for their exceptional software offerings, which played a pivotal role in enabling efficient data manipulation, interactive exploration, and robust machine learning experimentation. Their contributions have undeniably accelerated our research progress and enriched our analytical capabilities.

## Competing Interests

The authors declare no competing interests.

## Ethical and Informed Consent for Data Used

The study does not involve any human participants or personal data; hence, ethical approval and informed consent were not required. The data used in this research consists solely of linguistic datasets which are publicly available or generated for the purpose of this study.

## Data Availability and Access

The datasets generated and/or analyzed during the current study are available from the corresponding author on reasonable request. Additionally, the Persian homograph disambiguation dataset introduced in this study is made available for research purposes and can be accessed by contacting S.M. Ayyoubzadeh at smoein.ayyoubzadeh16@sharif.edu.